\documentclass{article}
\usepackage{spconf,amsmath,graphicx}
\usepackage{multirow}
\usepackage{booktabs}
\usepackage{mathrsfs}
\usepackage{amsfonts}
\usepackage{color}
\usepackage{url}

\title{Gaussian Prior Reinforcement Learning for\\ Nested Named Entity Recognition}
\name{ Yawen Yang$^{1}$, Xuming Hu$^{1}$, Fukun Ma$^{1}$, Shu'ang Li$^{1}$, Aiwei Liu$^{1}$, Lijie Wen$^{1*}$\thanks{$^{*}$Corresponding author. This work is supported by the National Key Research and Development Program of China (No.2019YFB1704003), the National Nature Science Foundation of China (No.62021002), Tsinghua BNRist and Beijing Key Laboratory of Industrial Big Data System and Application.} and Philip S. Yu$^{1,2}$}
\address
{
 $^{1}$School of Software, Tsinghua University, Beijing, China\\
 $^{2}$Department of Computer Science, University of Illinois at Chicago, Chicago, USA \\
 $^{1}$\url{yyw19@mails.tsinghua.edu.cn} $^{1}$\url{wenlj@tsinghua.edu.cn}
}
\begin{document}
%
\maketitle
\begin{abstract}
Named Entity Recognition (NER) is a well and widely studied task in natural language processing. Recently, the nested NER has attracted more attention since its practicality and difficulty. Existing works for nested NER ignore the recognition order and boundary position relation of nested entities. To address these issues, we propose a novel seq2seq model named \verb|GPRL|, which formulates the nested NER task as an entity triplet sequence generation process. \verb|GPRL| adopts the reinforcement learning method to generate entity triplets decoupling the entity order in gold labels and expects to learn a reasonable recognition order of entities via trial and error. Based on statistics of boundary distance for nested entities, \verb|GPRL| designs a Gaussian prior to represent the boundary distance distribution between nested entities and adjust the output probability distribution of nested boundary tokens. Experiments on three nested NER datasets demonstrate that \verb|GPRL| outperforms previous nested NER models. Source code will be available at \url{ https://github.com/THU-BPM/GPRLNER}.
\end{abstract}
\section{Introduction}
\label{sec:intro}
Named Entity Recognition (NER) aims to locate information entities and identify their corresponding categories, which is widely used in various downstream NLP tasks, including entity linking \cite{le-titov-2018-improving, ganea-hofmann-2017-deep} and relation extraction \cite{hu2020selfore, hu2021semi, liu2022hierarchical}. Nested NER refers to that a named entity contains one or more entities within it or is a part of them, such as ``South African" and ``South African scientists". Entity nested circumstances are common in different languages and domains. Traditional sequence labeling methods fail to handle nested NER since one token may belong to different entities.

Lots of efforts have been devoted to solve nested NER tasks effectively in recent years. Proposed methods could be mainly divided into sequence-based, span-based and hypergraph-based methods \cite{shen-etal-2021-locate}. The sequence-based methods \cite{strakova-etal-2019-neural,yan-etal-2021-unified-generative} treat nested NER as a sequence labeling or entity span sequence generation task. However, the former leads to a large growth of label categories and the sparse label distribution when combining BIO labels, while the latter faces exposure bias \cite{zhang-etal-2019-bridging} due to fixed order generation. The span-based methods \cite{sohrab-miwa-2018-deep, 2020Boundary} consider the nested NER as a classification task of all the candidate entity spans extracted from the input sentence. Obviously, this method brings high computational costs for numerous meaningless spans and ignores the dependency between nested entities. The hypergraph-based methods \cite{lu-roth:2015:EMNLP, katiyar-cardie-2018-nested} construct hypergraphs based on the nested structure and introduce graph neural network, which may suffer from the spurious structure of hypergraphs.

\begin{figure} 
  \centering
  \includegraphics[width=\linewidth]{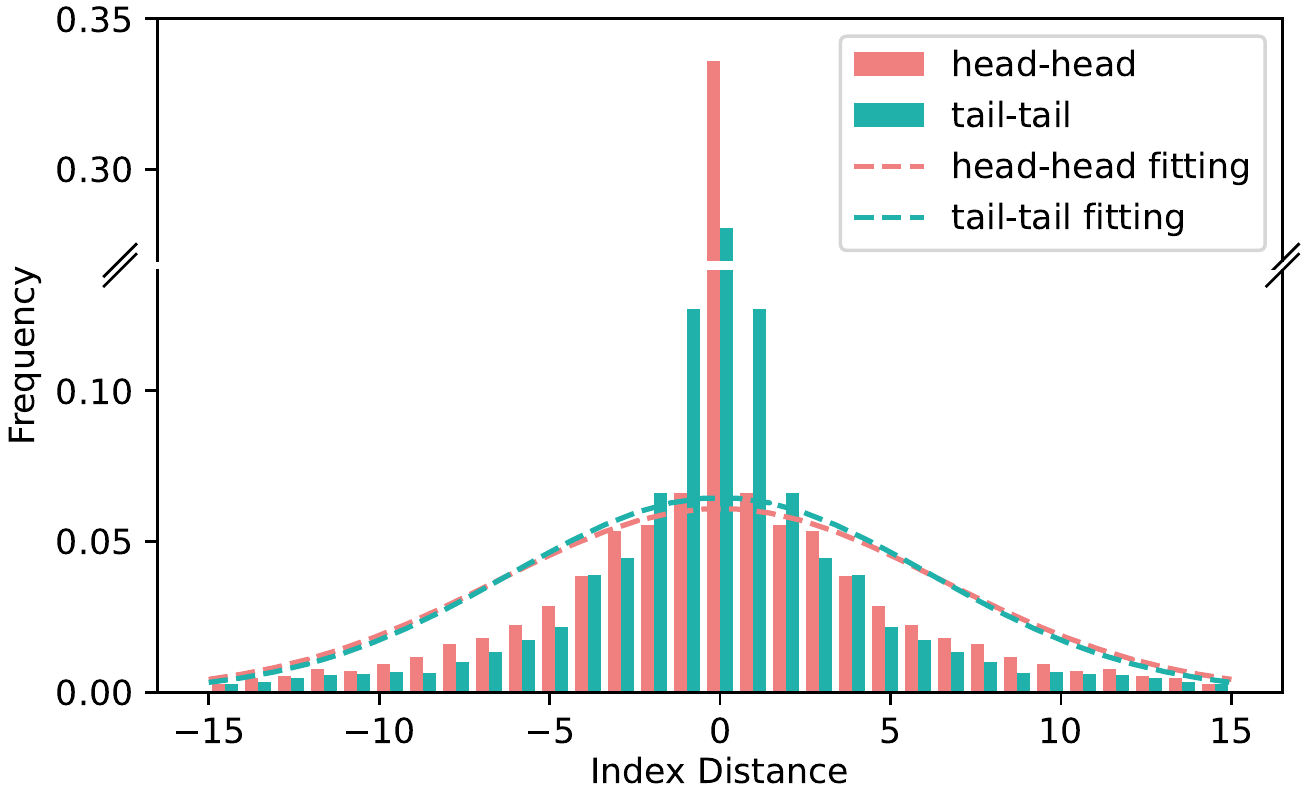}
  \caption{The boundary distance frequency distribution of all the nested entity pairs in ACE 2004 training data.}
  \label{fig:pic1_3}
  \vspace{-5mm}
  \setlength{\abovecaptionskip}{-5mm}
  \setlength{\belowcaptionskip}{-5mm}
\end{figure}

To reduce the above drawbacks, this paper further explores structural and semantic characteristics of nested entities, including (1) entity triplet recognition order and (2) boundary distance distribution of nested entities. For example,  it is intuitively reasonable that the identification of ``South African" may promote the prediction of ``South African scientists" entity. Inspired by methods with unfixed triplet order \cite{zeng-etal-2019-learning,tan2021sequencetoset}, we hope that the model can learn an optimal recognition order and get rid of the teacher forcing of gold triplet orders. In addition, we obtain boundary distance statistics of all the nested entity pairs in multiple corpora. From Figure \ref{fig:pic1_3}, we assume the boundary distance frequency of nested entity pairs can be naturally formulated into Gaussian distribution.



\begin{figure*} 
  \centering
  \includegraphics[width=0.9\linewidth]{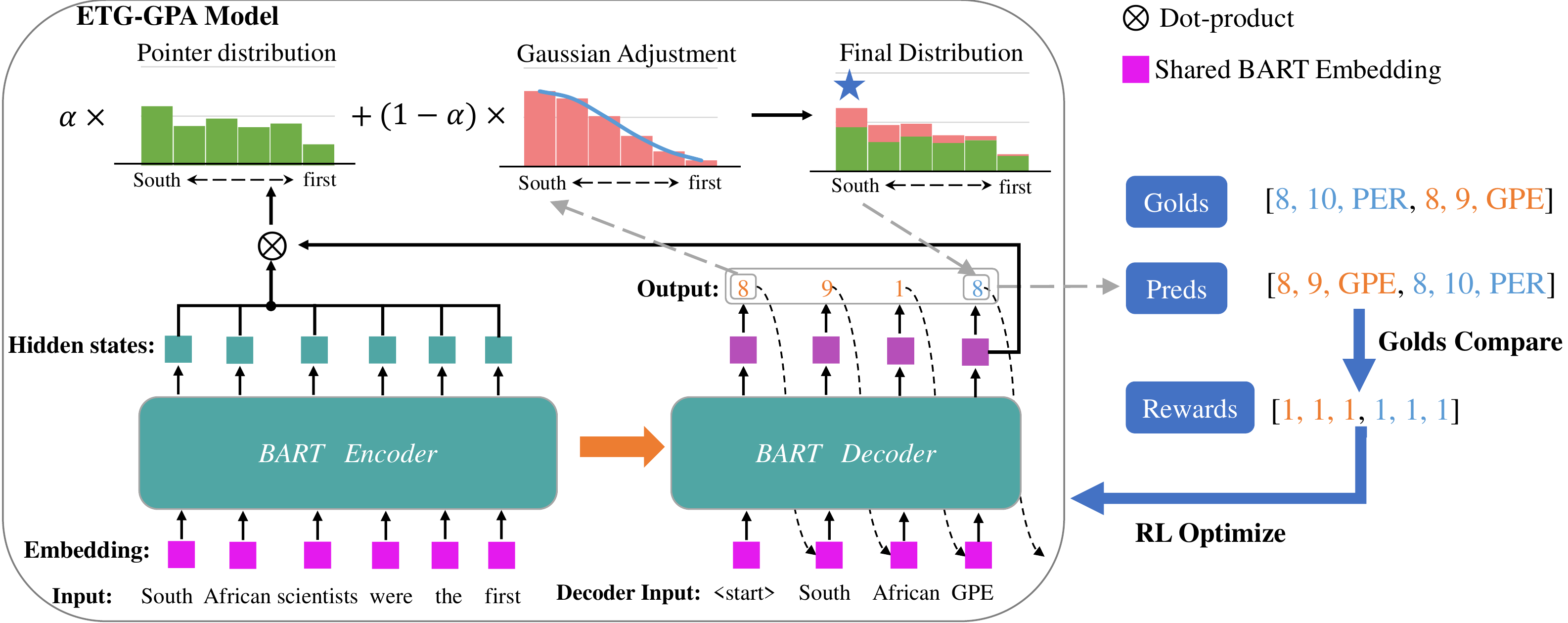}
  \caption{Overview of the proposed GPRL framework for nested NER. Since the input sentence is selected from ACE 2004 containing 7 entity categories, we conduct the 7 shift to the token index. Thus index 1-7 represents different entity categories.}
  \label{fig:pic1_5}
  \vspace{-5mm}
\end{figure*}

Armed with these observations, we propose a \textbf{G}aussian \textbf{P}rior \textbf{R}einforcement \textbf{L}earning framework named \verb|GPRL| to learn the entity recognition order and utilize the boundary position relation of nested entities. As shown in Figure \ref{fig:pic1_5}, \verb|GPRL| converts the nested NER task into an entity triplet sequence generation task and uses the pre-trained seq2seq model BART \cite{lewis-etal-2020-bart} with pointer mechanism \cite{NIPS2015-29921001} to generate the entity triplet sequence. In order to improve the ability of the model to recognize nested entities, we apply the Gaussian prior adjustment of nested entity pairs to the entity boundary probability predicted by pointer network. We further take the recognition order into consideration and model the entity triplet generation as a reinforcement learning (RL) process. Based on the generation quality of each triplet, we assign a reward to quantify model behavior and optimize the network through maximizing the rewards. Experimental results show that \verb|GPRL| achieves state-of-the-art on three public datasets and outperforms strong baselines on F1 score by 0.59\% on ACE 2004, 0.47\% on ACE 2005 and 0.06\% on GENIA.


\section{Proposed Model}
\label{sec:format}
\subsection{Entity Triplet Generator (ETG)}
The nested NER task could be formulated as follows: given an input sentence of $n$ tokens $X=[x_1, x_2, ..., x_n]$, we expect to identify all the named entities and obtain the target $Y=[s_1, e_1, t_1, ..., s_k, e_k, t_k]$, where $k$ is the total number of entities, $s_i, e_i(i \in [1, k])$ indicate the start and end index of the $i$-th entity respectively and $t_i$ represents the entity type. To get such entity span sequence, we select the pre-trained seq2seq model BART as the entity triplet generator. 

The encoder deals with the raw sentence $X$ to get the sequence representation $\mathbf{H}^e \in \mathbb R^{n \times d}$ as: 
\begin{equation}
\setlength{\abovedisplayskip}{3pt}
\setlength{\belowdisplayskip}{3pt}
\mathbf{H}^e=\text{BartEncoder}(X),
\end{equation}
where $n$ is the sentence length and $d$ is the hidden dimension.

The decoder aims to predict boundary index or entity type at each time step $P_t = P(\hat{y}_t | H^e, Y_{<t})$. However, the target label sequence cannot be put into the decoder network directly since it contains both token index and entity type. Following \cite{yan-etal-2021-unified-generative}, we also design an Index2Token mode to convert predicted indexes into practical tokens.
\begin{equation}
\setlength{\abovedisplayskip}{3pt}
\setlength{\belowdisplayskip}{3pt}
\label{index2token} \hat{y}_t = \begin{cases}
T_{y_t}, &if \  y_t \leq k \\
X_{y_t-k}, &if \  y_t > k, 
\end{cases}
\end{equation}
where $T$ is the entity type list with the length $k$.

After converting indexes into tokens in this way, we can put the encoder output and target sequence into the decoder module and obtain the last hidden state, which formulates as:
\begin{equation}
\setlength{\abovedisplayskip}{3pt}
\setlength{\belowdisplayskip}{3pt}
\mathbf{h}^d_t=\text{BartDecoder}(\mathbf{H}^e;\hat{Y}_{<t}), 
\end{equation}
where $\hat{Y}_{<t}=[\hat{y}_1, \hat{y}_2, ..., \hat{y}_{t-1}]$, representing the ground truth or the predicted span sequence for time step $t$.

Then we adopt the pointer mechanism to generate entity index probability distribution $P_t$ as:
\begin{align}
\mathbf{T}^d &= \text{BartTokenEmbed}(\mathbf{T}), \\
\mathbf{L}_t &= [\mathbf{T}^d \otimes \mathbf{h}^d_t; \mathbf{H}^e \otimes \mathbf{h}^d_t], \\
\label{label-score} P_t &= \text{Softmax}(\mathbf{L}_t),
\end{align}
where BartTokenEmbed means the token embeddings shared between encoder and decoder; $\mathbf{T}^d \in \mathbb R^{k \times d}$; $\mathbf{L}_t \in \mathbb R^{t \times (k+n)}$; $\otimes$ means the dot product of multi-dimension vectors and $[\cdot;\cdot]$ concats the two vectors in the first dimension.

\subsection{Gaussian Prior Adjustment (GPA)}
Through pointer network, we can copy words from input sequence instead of searching the whole output vocabulary. In Eq.\ref{label-score}, the $\mathbf{L}_t$ vector works as the attention distribution, concentrating on the token with the maximum score in the original sequence. As discussed above, the boundary distance between nested entity pairs roughly obeys the Gaussian distribution, which indicates that the closer distance between the start (or end) positions of nested entity pairs has higher probability. Next, we adopt this characteristic to adjust the $P_t$ distribution, enhancing the recognition of nested entity boundaries.

In the process of generating entity triplet sequence $Y$, once getting the $j$-th triplet $(s_j, e_j, t_j)$, we search the already generated triplet list forward. If the $i$-th triplet has the nested relationship with the $j$-th triplet ($i<j$), we utilize boundary position of the previous nearest $i$-th entity to adjust the boundary distribution of current $j$-th entity by Gaussian prior. For simplicity, we choose the standard normal distribution with $1/(2\pi)$ variance whose probability density function is $\phi(d)=e^{-\pi d^2}$, $d$ is the distance between token positions. Then we replace the coefficient $\pi$ with a learnable parameter and compute Gaussian distribution by the relative distance between token position of input sentence and  boundary position of the $i$-th entity.
\begin{align}
P_{gass\_h}(m) &= \frac{e^{- \lambda (m-s_i)^2}}{\sum_{b=1}^{n} e^{- \lambda (b-s_i)^2}}, \\
P_{gass\_t}(m) &= \frac{e^{- \mu (m-e_i)^2}}{\sum_{b=1}^{n} e^{- \mu (b-e_i)^2}},
\end{align}
where $m \in [1,n]$ is the token index, $\lambda$ and $\mu$ are learnable parameters. Then the final probability distribution of the $j$-th entity boundary is:
\begin{align}
P_{s_j} &= \alpha * P_{head} + (1-\alpha) * P_{gass\_h}, \\
P_{e_j} &= \alpha * P_{tail} + (1-\alpha) * P_{gass\_t},
\end{align}
where $\alpha$ is the hyper-parameter, $P_{head}$ and $P_{tail}$ represent the probability distribution of the head and tail token after the pointer network in Eq.\ref{label-score} respectively.

\subsection{Entity Order Reinforcement Learning (EORL)}
In conventional seq2seq framework, entity triplets need to be generated one by one in a fixed order, which ignores the internal dependency between nested entities and suffers from exposure bias. To learn the recognition order of nested entities, we propose the Entity Order Reinforcement Learning (\verb|EORL|) component, which optimizes the \verb|ETG-GPA| model with REINFORCE algorithm \cite{Williams1992Simple}. The loop in Figure \ref{fig:pic1_5} represents a RL episode, where the \verb|ETG-GPA| model reads in an input sentence and generates the entity triplet sequence freely. We calculate the rewards based on the generation quality of each triple and optimize the model through trial and error. Now we explain the main ingredients of \verb|EORL| in detail.

\textbf{State:} Given the raw sentence of $n$ tokens $X=[x_1, x_2, ...\\, x_n]$ and the already generated entity index sequence $\hat{Y}_{<t}$, we use $s_t$ to denote the optimization state at time step $t$.
\begin{equation}
\setlength{\abovedisplayskip}{3pt}
\setlength{\belowdisplayskip}{3pt}
s_t=(\hat{Y}_{<t}, X, \theta)
\end{equation}
where $\theta$ represents parameters of \verb|ETG|-\verb|GPA| model.

\textbf{Policy:} We use the policy gradient algorithm in reinforcement learning, aiming to train a policy network that can generate entity triplets without fixed order limitation. The policy network is parameterized by the \verb|ETG|-\verb|GPA| model $f_\theta$. 

\textbf{Action:} The action is the boundary index or entity type $\hat{y}_t$ sampled from output probability distribution of the \verb|ETG|-\verb|GPA| model at time step $t$, which contains two cases as Eq.\ref{index2token}.

\textbf{Reward:} The reward is used to signal the generation quality of entity triplets and guide the model training, which plays an important role in RL process \cite{zeng-etal-2019-learning,hu2021gradient}. Due to the nature of seq2seq model, we cannot calculate the reward of each step directly since it is difficult to evaluate the quality of a single token. Therefore, we assign the reward every three time steps when a new triplet has been generated. One high-quality triplet needs to meet: (1) Its boundary and category are both correctly predicted; (2) It is not the same as any triplet previously generated. When we obtain such good triplet after three steps, we give the reward $1$ to each of these steps. Meanwhile, we define the empty triplet as entity triplets containing ending index or correct boundary and assign the reward 0.5 to each corresponding step. In other circumstances, the reward is highly close to 0. Assuming the \verb|ETG-GPA| model generates a new entity triplet $F_i$ in three steps ($t_i,t_{i+1},t_{i+2}$), the gold triplet list is $G$, the already generated triplet list is $V$, the reward assignment can be formulated as:
\begin{equation}
\setlength{\abovedisplayskip}{3pt}
\setlength{\belowdisplayskip}{3pt}
r_i=r_{i+1}=r_{i+2}=\begin{cases}
1, &\text{if} \ F_i \in G \ \text{and} \ F_i \notin V \\
0.5, &\text{if} \ F_i \ \text{is \ empty \ triplet} \\
0, &\text{otherwise}\\
\end{cases}
\end{equation}
\\
\textbf{Reinforcement Learning Loss} \\
According to the REINFORCE algorithm and policy gradient mechanism, we optimize the \verb|ETG|-\verb|GPA| network by \verb|EORL| through the following reinforcement learning loss:  
\begin{equation}
\setlength{\abovedisplayskip}{3pt}
\setlength{\belowdisplayskip}{3pt}
\mathcal{L}(\theta) = \frac{1}{T}\sum_{t=1}^{T}{loss(f_\theta(X, Y_{<t}), \hat{y}_t)*r_t},
\end{equation}
where $loss$ is the cross entropy loss function, $r_t$ is the reward, $f_\theta$ represents the policy network that generates the entity index and $\hat{y}_t$ is the sampled action from the index probability distribution. We minimize $\mathcal{L}(\theta)$ to optimize the $f_\theta$ model.

\section{Experiments}
\label{sec:pagestyle}

\subsection{Datasets and Settings}
To evaluate the proposed model, we conduct main experiments on three public nested NER datasets: ACE 2004 \cite{doddington-etal-2004-automatic}, ACE 2005 \cite{2006ACE} and GENIA \cite{10.1093/bioinformatics/btg1023}. For the evaluation metrics, we employ the span-level F1 score.


For three NER datasets, we add entity tags (``Person'', ``Location'', etc.) as special tokens to the BART Tokenizer. Towards model components, we use the pre-trained BART$_{large}$ as Entity Triple Generator. In Gaussian Prior Adjustment, we select weight coefficient $\alpha$ in a range of [0.7, 0.9] to avoid excessive interference brought by the prior probability. During model training, we first pre-train \verb|ETG-GPA| model with cross-entropy loss to achieve over 90\% of the best supervised learning performance, then train the model with reinforcement learning loss. By this way, we can improve the convergence rate and get stable running results. We use the AdamW optimizer with the learning rate set to 5e-5 for supervised learning, 5e-6 for RL learning respectively.

\begin{table}
\centering
\resizebox{\linewidth}{!}{ 
\begingroup
\renewcommand{\arraystretch}{0.8} 
\begin{tabular}{lccc}
\toprule
Model & ACE 2004 & ACE 2005 & GENIA \\
\midrule
\textbf{Sequence-based Methods} &  &  & \\
Seq2Seq \cite{strakova-etal-2019-neural} & 84.40 & 84.33 & 78.31\\
Pyramid \cite{wang-etal-2020-pyramid} & 86.28 & 84.66 & 79.19\\
Sequence-To-Set$^\ast$ \cite{tan2021sequencetoset} & 87.26 & 87.05 & 80.44\\
BartNER$^\dagger$ \cite{yan-etal-2021-unified-generative} & 86.84 & 84.74 & 79.23\\
\midrule
\textbf{Others} &  &  & \\
Biaffine$^\ast$ \cite{yu-etal-2020-named} & 86.70 & 85.40 & 80.50\\
Locate-and-Label$^\ast$ \cite{shen-etal-2021-locate} & 87.41 & 86.67 & 80.54\\
Triaffine$^\ast$ \cite{yuan-etal-2022-fusing} & 87.40 & 86.82 & 81.23 \\
W2NER$^\ast$ \cite{li2021unified} & 87.52 & 86.79 & 81.39 \\
Lexicalized-Parsing$^\ast$ \cite{lou-etal-2022-nested} & 87.90 & 86.91 & 78.44 \\
\midrule
\textbf{Ours GPRL}$^\dagger$ & \textbf{88.49} & \textbf{87.52} & \textbf{81.45}\\
\ $-GPA$ & 87.62 & 86.99 & 80.51\\
\ $-EORL$ & 87.06 & 86.27 & 79.93\\
\bottomrule
\end{tabular}
\endgroup
}
\caption{\label{main-result}
Overall performances on nested NER datasets. BERT is the default encoder, $\ast$ represents BERT$_{large}$ and $\dagger$ represents BART$_{large}$ encoder. We report the average F1 results with 5 runs of training and testing.
}
\vspace{-4mm}
\end{table}

\subsection{Results and Comparisons}
Since the proposed model bases on seq2seq structure, we first compare it with several outstanding sequence-based methods: \textbf{Seq2Seq} \cite{strakova-etal-2019-neural}, \textbf{Pyramid} \cite{wang-etal-2020-pyramid}, \textbf{Sequence-To-Set} \cite{tan2021sequencetoset} and \textbf{BartNER} \cite{yan-etal-2021-unified-generative}. Then we introduce other state-of-the-art models for further comparison: \textbf{Biaffine} \cite{yu-etal-2020-named}, \textbf{Locate-and-Label} \cite{shen-etal-2021-locate}, \textbf{Triaffine} \cite{yuan-etal-2022-fusing}, \textbf{W2NER} \cite{li2021unified}, \textbf{Lexicalized-Parsing} \cite{lou-etal-2022-nested}.

The overall performances of our proposed model on nested datasets are shown in Table \ref{main-result}. We could observe that \verb|GPRL| outperforms all the baseline models consistently on each dataset. To be specific, compared with previous SOTA models, \verb|GPRL| achieves +0.59\% F1 higher on ACE 2004 and +0.47\% F1 higher on ACE 2005. Meanwhile, \verb|GPRL| achieves the comparable result (+0.06\% F1 higher) with W2NER \cite{li2021unified} on GENIA dataset. Experimental results demonstrate the effectiveness of \verb|GPRL| model to identify nested entities.


\subsection{Ablation Study and Further Analysis}
To further prove the effectiveness of several modules in \verb|GPRL| model, we conduct necessary ablation experiments. As shown in Table \ref{main-result}, when we remove the \verb|GPA| component, the F1 score decreases by 0.87\% on ACE 2004, 0.53\% on ACE 2005 and 0.94\% on GENIA. Since the Gaussian prior is designed directly from nested entity pairs without considering flat entities, we believe \verb|GPA| helps to recognize nested entity boundaries. If we delete the \verb|EORL| module and optimize \verb|ETG-GPA| network only with cross entropy loss, the F1 performance decreases by 1.43\% on ACE 2004, 1.25\% on ACE 2005 and 1.52\% on GENIA, respectively. This indicates that the reinforcement learning process can learn the proper recognition order of entities and weaken the training inference gap through negative samples for seq2seq models.

\begin{figure} 
  \centering
  \includegraphics[width=0.90\linewidth]{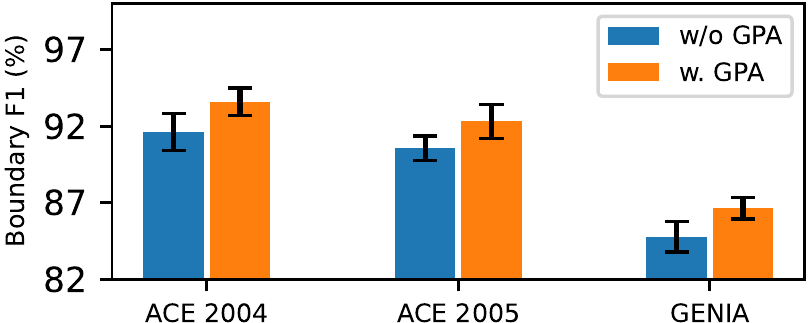}
  \caption{Comparisons of F1 score on entity boundary detection.}
  \label{fig:pic3_6}
  \vspace{-2mm}
\end{figure}

The main purpose of \verb|GPA| is to increase the attention score of neighboring tokens and decay the importance of distant ones for previous nested boundary. To further prove its effectiveness, we compare the change of F1 on entity boundary detection. As shown in Figure \ref{fig:pic3_6}, the F1 with \verb|GPA| is obviously higher on three datasets, indicating \verb|GPA| helps the model locate nested boundaries and redress incorrect predictions.

We finally study cases where entity locating and classifying are improved with \verb|EORL|. From Table \ref{rl-case}, the \verb|EORL| solves the problems of less labeling and error identification effectively. We also discover that \verb|EORL| tends to identify internal entities first in nested situation and learns a proper generation order of multiple entities. Model with \verb|EORL| can capture the dependencies and interactions between nested entities by generating entity triplets in flexible orders. 

\begin{table}
\centering
\resizebox{\linewidth}{!}{
\begin{tabular}{l}
\toprule
(A (U.S.) tourist) was detained after photographing a riot in (the\\ province of (Irian Jaya)).\\
\textbf{Label}: [0, 2, PER, 1, 1, GPE, 10, 14, GPE, 13, 14, GPE] \\
w/o \textbf{EORL}: [0, 2, PER, 1, 1, GPE, 13, 14, GPE] \\
w. \textbf{EORL}: [0, 2, PER, 1, 1, GPE, 13, 14, GPE, \textcolor{blue}{10, 14, GPE}] \\
\\
The deadly disease attacks ((African) villages) and kills (up to \\ 90\% of (those infected)).\\
\textbf{Label}: [4, 4, GPE, 4, 5, GPE, 8, 13, PER, 12, 13, PER] \\
w/o \textbf{EORL}: [4, 4, GPE, 4, 5, GPE, \textcolor{red}{10, 13, PER}, 12, 13, PER] \\
w. \textbf{EORL}:  [4, 4, GPE, 4, 5, GPE, 12, 13, PER, \textcolor{blue}{8, 13, PER}]\\
\bottomrule
\end{tabular}
}
\caption{\label{rl-case}
Predictions with/without EORL, where \textcolor{red}{red} represents incorrect boundary location, \textcolor{blue}{blue} represents change of entity recognition order and the number is token index.    
}
\vspace{-5mm}
\end{table}

\section{Conclusion}
\label{sec:typestyle}

In this paper, we propose a novel reinforcement learning seq2seq model \verb|GPRL| for nested NER. We explore the boundary distance distribution of nested entity pairs, which is formulated as a Gaussian prior to adjust the pointer scores and helps to identify nested boundaries. Different from conventional seq2seq model, we make the model generate entity triplets freely through reinforcement learning process so that it can actually learn the recognition order. Experiments on three nested NER datasets show \verb|GPRL| achieves SOTA performance over strong baselines.

\ninept

\bibliographystyle{IEEEbib}
\bibliography{strings,refs}

\end{document}